\documentclass[sigconf]{acmart}

\AtBeginDocument{%
  \providecommand\BibTeX{{%
    \normalfont B\kern-0.5em{\scshape i\kern-0.25em b}\kern-0.8em\TeX}}}

\copyrightyear{2021}
\acmYear{2021}
\setcopyright{acmcopyright}\acmConference[MM'21]{Proceedings of the 29th ACM International Conference on Multimedia}{October 20--24, 2021}{Virtual Event, China} \acmBooktitle{Proceedings of the 29th ACM International Conference on Multimedia (MM'21), October 20--24, 2021, Virtual Event, China}
\acmPrice{15.00}
\acmDOI{10.1145/3474085.3475251}
\acmISBN{978-1-4503-8651-7/21/10}

\usepackage{epsfig}
\usepackage{graphicx}
\usepackage{amsmath}

\usepackage{amssymb}
\usepackage[ruled,vlined]{algorithm2e}
\usepackage{multirow}
\usepackage{makecell}
\usepackage{url}
\usepackage{colortbl}
\usepackage{booktabs}
\usepackage{caption}
\usepackage{subcaption}
\usepackage{libertine}
\usepackage[export]{adjustbox}
\usepackage{colortbl}
\definecolor{mygray1}{gray}{.9}
\definecolor{mygray2}{gray}{.7}

\begin{document}
\title{ROSITA: Enhancing Vision-and-Language Semantic Alignments via Cross- and Intra-modal Knowledge Integration}

\author{
    Yuhao Cui${^{1}}$, Zhou Yu${^{1*}}$, Chunqi Wang${^2}$, Zhongzhou Zhao${^2}$, Ji Zhang${^2}$, Meng Wang${^3}$, Jun Yu${^1}$
}
\thanks{*Zhou Yu is the corresponding author.\\Work was done when Yuhao Cui was an intern at Alibaba Group.\\Code available at \url{https://github.com/MILVLG/rosita}}
\affiliation{
      \institution{${^1}$School of Computer Science and Technology, Hangzhou Dianzi University, China}
    }
\affiliation{
      \institution{${^2}$Alibaba Group, China}
    }
\affiliation{
      \institution{${^3}$School of Computer Science and Information Engineering, Hefei University of Technology, China}
    }
\email{{cuiyh, yuz, yujun}@hdu.edu.cn, {shiyan.wcq, zhongzhou.zhao, zj122146}@alibaba-inc.com, eric.mengwang@gmail.com}
\renewcommand{\authors}{Yuhao Cui, Zhou Yu, Chunqi Wang, Zhongzhou Zhao, Ji Zhang, Meng Wang, Jun Yu}



\begin{abstract}
Vision-and-language pretraining  (VLP)  aims to learn generic multimodal representations from massive image-text pairs. While various successful attempts have been proposed, learning fine-grained semantic alignments between image-text pairs plays a key role in their approaches. Nevertheless, most existing VLP approaches have not fully utilized the intrinsic knowledge within the image-text pairs, which limits the effectiveness of the learned alignments and further restricts the performance of their models. To this end, we introduce a new VLP method called ROSITA, which integrates the c\textbf{\underline{ROS}}s- and \textbf{\underline{I}}n\textbf{\underline{T}}r\textbf{\underline{A}}-modal knowledge in a unified scene graph to enhance the semantic alignments. Specifically, we introduce a novel structural knowledge masking (SKM) strategy to use the scene graph structure as a priori to perform masked language (region) modeling, which enhances the semantic alignments by eliminating the interference information within and across modalities. Extensive ablation studies and comprehensive analysis verifies the effectiveness of ROSITA in semantic alignments. Pretrained with both in-domain and out-of-domain datasets, ROSITA significantly outperforms existing state-of-the-art VLP methods on three typical vision-and-language tasks over six benchmark datasets.
\end{abstract}



\maketitle
\fancyhead{}


\section{Introduction}
Motivated by the success of the \emph{pretrain-then-finetune} paradigm of BERT in natural language understanding \cite{devlin2019bert}, there has been an increasing interest in developing vision-and-language pretraining (VLP) models \cite{lu2019vilbert,tan2019lxmert,chen2020uniter,li2020oscar} to address a wide range of vision-and-language (V+L) tasks. In particular, these approaches first pretrain transformer-based models on large image-text corpus to learn task-agnostic representations, and then finetune the models on downstream V+L tasks, \emph{e.g.}, visual question answering \cite{yu2017mfb,yu2018beyond}, image text retrieval \cite{plummer2015flickr30k,lee2018stacked}, and referring expression comprehension \cite{kazemzadeh2014referitgame,yu2018rethining}. Compared to earlier methods that are only adapted to one V+L task \cite{yu2019deep,yu2018mattnet,yu2020deep}, VLP models is generalizable across multiple tasks and also achieves significantly better performance on respective tasks.

\begin{figure}
\begin{center}
\includegraphics[width=0.47\textwidth]{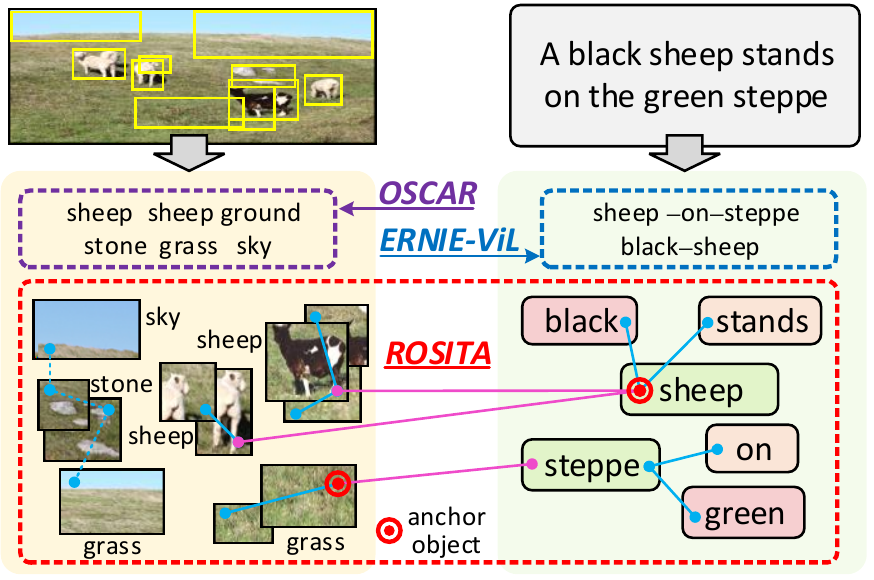}
\vspace{-5pt}
\caption{Schematic of the knowledge integration strategies of three VLP methods, \emph{i.e.}, OSCAR \cite{li2020oscar}, ERNIE-ViL \cite{yu2020ernievil}, and our ROSITA. OSCAR and ERNIE-ViL only exploit the intra-modal knowledge from the image and text modalities, respectively. In contrast, ROSITA simultaneously encodes the cross-modal knowledge (pink line) and intra-modal knowledge (blue line) in a unified scene graph centered at specific anchor objects, which is used to enhance the learning of fine-grained semantic alignments across modalities.}
\label{fig:example}
\end{center}
\vspace{-10pt}
\end{figure}

Learning \emph{fine-grained} semantic alignments between image regions and text words plays a key role in V+L tasks. However, manually annotating such dense alignment between regions and words is expensive and is unrealistic under the large-scale scenario. Therefore, most existing VLP approaches \cite{chen2020uniter,lu2019vilbert,li2019visualbert} use a weakly-supervised learning strategy to model the alignments implicitly. Taking the image regions and text words as inputs, they adopt multi-layer Transformers \cite{vaswani2017attention} as their backbones to learn \emph{fine-grained} semantic alignments from \emph{coarse-grained} image-text matching supervision. Moreover, the interference within and across modalities makes the learning of semantic alignments even more challenging.

To facilitate the learning of semantic alignments, two recent VLP approaches OSCAR \cite{li2020oscar} and ERNIE-ViL \cite{yu2020ernievil} introduce extra knowledge in different ways. Specifically, OSCAR additionally extracts the predicted region tags from images and uses these tags as anchor points to align with text words implicitly. ERNIE-ViL explicitly constructs a scene graph from text and puts more emphasis on the keywords (\emph{e.g.}, objects along with their attributes and relations) in the scene graph in its pretraining objectives. In terms of  knowledge source, both of them use the \emph{intra-modal} knowledge from a single modality to enhance the semantic alignments: OSCAR models the intra-modal knowledge in the image modality while ERNIE-ViL models the intra-modal knowledge in the text modality. The success of the two methods above raises a question: \emph{Is it possible to utilize the intra-modal knowledge from both modalities along with the cross-modal knowledge to further enhance the semantic alignments?}

In this paper, we present a new VLP method called ROSITA, which encodes the c\textbf{\underline{ROS}}s- and \textbf{\underline{I}}n\textbf{\underline{T}}r\textbf{\underline{A}}-modal knowledge simultaneously in a unified scene graph. As shown in Figure \ref{fig:example}, the graph consists of a set of knowledge entries, where each entry corresponds to an \emph{anchor object} along with its associated cross- and intra-modal knowledge. The intra-modal knowledge refers to the relationships between the anchor object and its intra-modal contexts (\emph{e.g.}, spatially related regions or contextually related words). The cross-modal knowledge corresponds to the relationships between the anchor object and its semantically similar objects from the opposite modality (\emph{e.g.}, the region predicted as  ``\emph{grass}'' and the word ``\emph{steppe}'').

Although we have obtained a set of knowledge entries, how to effectively use them to enhance semantic alignments is still nontrivial. We propose a novel \emph{structural knowledge masking} (SKM) strategy that can be seamlessly integrated with the masked language (region) modeling tasks, which are commonly used in existing VLP methods \cite{chen2020uniter,lu2019vilbert}. In principle, SKM determinately masks the anchor object while selectively masking its cross- and intra-modal contexts in a knowledge entry. This strategy effectively eliminates the interference information within and across modalities and enhances the semantic alignments by enforcing the model to acquire accurate information from the \emph{opposite} modality.


The contributions of this work are three-fold:
\begin{enumerate}
  \item We present a new VLP method ROSITA, which incorporates cross- and intra-modal knowledge simultaneously to enhance the semantic alignments across different modalities.
  \item We introduce a novel structural knowledge masking strategy to use the scene graph structure as a priori to be integrated with the commonly used masked language (region) modeling tasks in existing VLP methods.
  \item We achieve the best results on three typical V+L tasks over six benchmark datasets, outperforming existing state-of-the-art VLP methods.
\end{enumerate}

\section{Related work}
We briefly review previous studies on unimodal pretraining and vision-and-language pretraining, especially those studies on knowledge enhanced pretraining.
\vspace{5pt}
\\
\noindent\textbf{Unimodal Pretraining.} The pretraining technique has been widely used in computer vision (CV) tasks. Deep convolutional neural networks like VGGNet \cite{simonyan2014very} or ResNet \cite{he2016deep} pretrained on ImageNet can well generalize to various downstream tasks  \cite{ren2015faster,long2015fully,he2017mask}. In contrast to CV tasks, the popularization of pretraining in the natural language processing (NLP) community is relatively late. Based on the multi-layer Transformer architecture \cite{vaswani2017attention}, many famous pretraining approaches (\emph{e.g.}, BERT \cite{devlin2019bert}, GPT \cite{radford2018improving}, and XLNet \cite{yang2019xlnet}) have been put forward. Different from the supervised pretraining paradigm in CV tasks, the pretraining paradigm in NLP tasks is \emph{self-supervised} that aims to train a model to predict words based on their contexts without introducing human annotations. In particular, BERT introduces a novel masking language modeling (MLM) task that randomly masks the input words and predicts these masked words based on their contexts. This MLM strategy is naturally inherited by the VLP methods.
\vspace{5pt}
\\
\noindent\textbf{Vision-and-Language Pretraining (VLP).} Different from the purely self-supervised paradigm in NLP tasks, VLP models are pretrained on large-scale paired image-text corpus, \emph{e.g.},  image captioning datasets like \cite{chen2015microsoft,sharma2018conceptual,ordonez2011im2text}. Mirroring the success of BERT, recent studies naturally extend its framework to the vision-and-language domain to pretrain VLP models for a wide range of V+L tasks \cite{chen2020uniter,li2020oscar,zhang2020devlbert,lu2019vilbert,tan2019lxmert,yu2020ernievil,huang2020pixel}. ViLBERT \cite{lu2019vilbert} and LXMERT \cite{tan2019lxmert} are two pioneering works in this field, where the two-stream architectures are adopted to encode the image features and textual features with two separate Transformers and then perform multimodal fusion via a third Transformer. Recent works tend to use the single-stream architectures, where the multimodal features are directly fused using one Transformer \cite{chen2020uniter,li2019visualbert,li2020unicoder,su2019vl}. Moreover, other techniques like knowledge integration \cite{yu2020ernievil,li2020oscar}, multilingual enhancement \cite{zhou2021uc2}, contrastive learning \cite{li2020unimo}, and adversarial training \cite{gan2020large} are introduced to further improve the performance of the pretrained models.
\vspace{5pt}
\\
\noindent\textbf{Knowledge-Enhanced Pretraining.} Incorporating prior knowledge (\emph{e.g.}, external knowledge graph) to enhance model pretraining has been investigated earlier by two ERNIE methods \cite{sun2019ernie,zhang2019ernie} and widely explored in recent years \cite{liu2020kbert,wang2021kepler,wang2020k}. The introduced prior knowledge enables the model to better understand the syntactic and semantic structure of the text, thus facilitating model pretraining by an improved structural MLM task. In the VLP task, prior knowledge can be acquired from both the image and text modalities. ERNIE-ViL constructs a scene graph from text and puts more emphasis on the discovered keywords \cite{yu2020ernievil}. OSCAR exploits the predicted tags of image regions to enhance the semantic alignment across the two modalities \cite{li2020oscar}. A concurrent work UC2 utilizes off-the-shelf machine translation model to construct aligned multilingual dataset for texts and regard this extra information as prior knowledge to enhance the learning of cross-modal semantic alignment \cite{zhou2021uc2}. Despite the success of these knowledge-enhanced VLP methods, they only utilize the
intra-modal knowledge from a single modality, which restricts their effectiveness in learning semantic alignments.

To the best of our knowledge, our ROSITA is the first VLP method to integrate the cross-modal and intra-modal knowledge simultaneously in order to enhance the learning of semantic alignments across different modalities.
\begin{figure*}
\begin{center}
\includegraphics[width=0.98\textwidth]{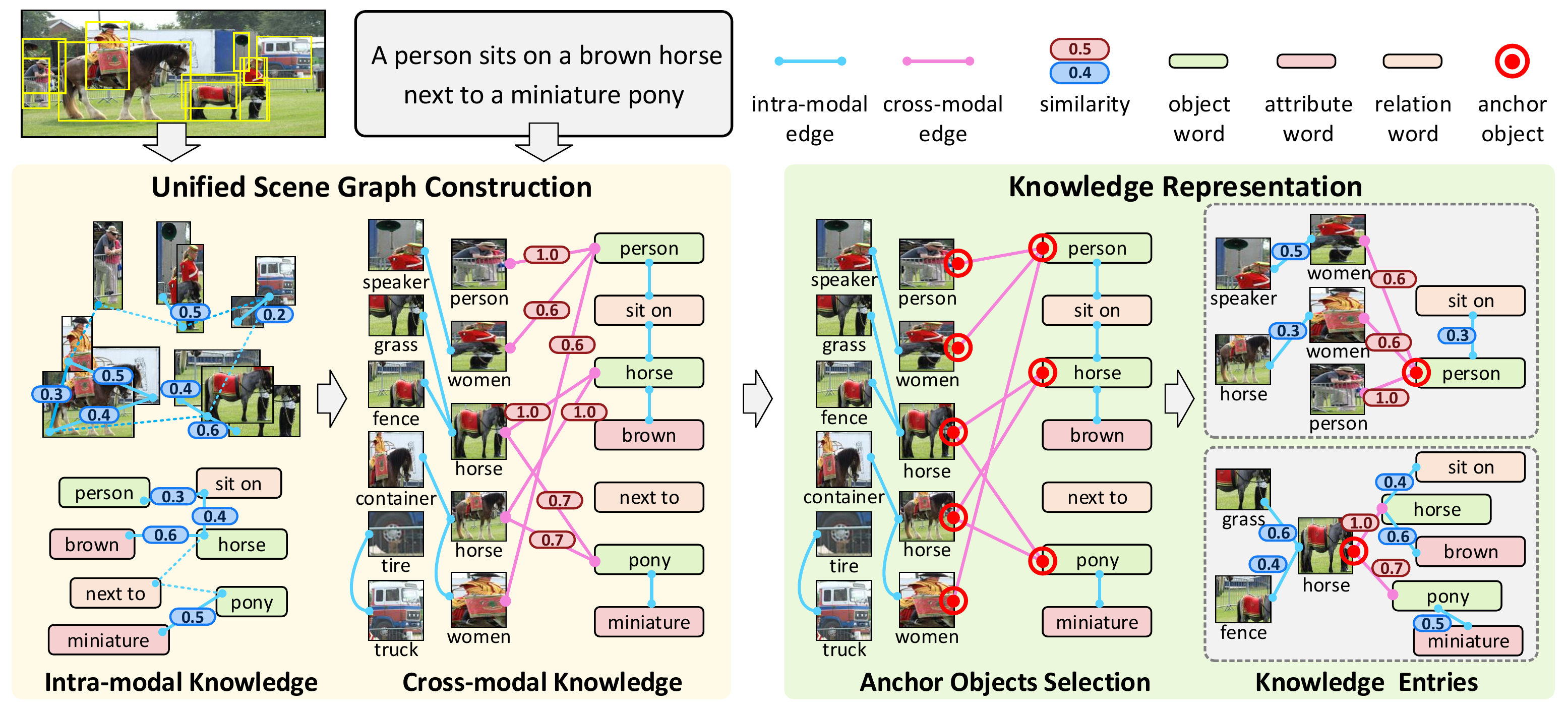}
\vspace{-5pt}
\caption{The flowchart of knowledge extraction given an image-text pair. It consists of two main stages, namely the unified scene graph construction and knowledge representation.}
\label{fig:rosita}
\end{center}
\vspace{-10pt}
\end{figure*}
\vspace{5pt}
\section{Knowledge Extraction}
In this section, we introduce the procedure of extracting knowledge entries from an image-text pair. We first construct a unified graph to model the intra- and cross-modal knowledge from an image-text pair. On top of the established graph, we select anchor objects to obtain a set of knowledge entries. The process of knowledge extraction is illustrated in Figure \ref{fig:rosita}.
\vspace{5pt}
\\
\noindent\textbf{Unified Scene Graph Construction.}
Given an image-text pair, we resort to a unified scene graph structure $G=<V, E, S>$ to encode its intra- and cross-modal knowledge simultaneously \cite{yu2014discriminative}. The vertex set $V$ includes the words and regions from the text and image, respectively. The edge set $E$ and similarity set $S$ contain pairwise relationships and their corresponding similarities between vertices (\emph{i.e.}, edge weights), respectively.

The intra-modal knowledge within the image and text are first represented as an image scene graph and a text scene graph, respectively. For the image scene graph, regions extracted from a pretrained object detector are considered as the vertices in $V$. Inspired by \cite{li2019relation,yao2018exploring,kant2020spatially}, we calculate the similarity between each paired regions by their Intersection over Union (IoU) score. The region pairs with IoU scores larger than zero are considered to have edges in $E$ and their IoU scores are regarded as their similarities in $S$. For the text graph, we use an off-the-shelf scene graph parser provided by \cite{anderson2016spice} to obtain a text scene graph from a text. The text scene graph explicitly encodes the keywords of objects, attributes, and relations found in the text while discarding the rest of uninformative words. These mentioned keywords in the scene graph are regarded as the vertices in $V$. The word-word relationships in the scene graph (\emph{i.e.}, object-attribute or object-relation) correspond to the edges in $E$. The similarity between two vertices is the co-occurrence frequencies of the referred object-attribute (or object-relation) pair calculated on the whole dataset. Since the similarity distributions of the image and text modalities may vary widely, we normalize the similarities within each modality, respectively.

As we have modeled the intra-modal knowledge in the graph, we further integrate cross-modal knowledge to align the image regions to their semantically related words. Since such cross-modal alignment supervision is not available, we establish \emph{pseudo} semantic alignments between region-word pairs as follows. For the image regions, the predicted region tags are aligned to the object words with respect to their semantic similarities on words. We adopt a pretrained word embedding model \cite{pennington2014glove} to calculate the pairwise similarities between object tags and object words\footnote{We have tried to establish more fine-grained alignments to include the attribute words. However, the predicted attributes from image regions are too diverse that often fail to match the attribute words in the text.}. We set a minimum confidence threshold of 0.5 to the similarity scores to make a trade-off between precision and recall. The region-word pairs surpass the threshold will form cross-modal edges in $E$ and their corresponding scores represent the similarities in $S$.
\vspace{5pt}
\\
\noindent\textbf{Knowledge Representation.} Based on the constructed unified scene graph $G$, we illustrate the procedure of extracting knowledge entries from the scene graph in detail. Note that each knowledge entry is associated with an anchor object, we first select all possible anchor objects from the graph. We define an anchor object as the vertex (an image region or a text word) in the graph that is referred to by at least one cross-modal edge. Since the attribute and relation words are not directly connected to any image region, they cannot be anchor objects according to our definition.

After obtaining the anchor objects, we integrate the intra-modal knowledge and cross-modal knowledge in $G$ to obtain a knowledge entry. Given an anchor object $v\in V$, its corresponding knowledge entry is represented as a subgraph $g(v) \subseteq G$ and is obtained by the union of three subgraphs of $G$ as follows:
\begin{equation}\label{eq:rosita}
g(v)=G_\mathrm{cross}(v)\cup G_\mathrm{intra}(v)\cup G_\mathrm{intra}(G_\mathrm{cross}(v))
\end{equation}
where $G_\mathrm{cross}(v)$ contains the relationships between $v$ and its {directly connected} contexts by cross-modal edges. $G_\mathrm{intra}(v)$ models the relationships between $v$ and its directly connected contexts by intra-modal edges. $G_\mathrm{intra}(G_\mathrm{cross}(v))$ includes the relationships between the vertices in $G_\mathrm{cross}(v)$ and their corresponding intra-modal contexts. It is worth noting that the anchor object $v$ can reach every vertex in $g(v)$ within two hops.

\section{The ROSITA Framework}
\begin{figure}
\begin{center}
\includegraphics[width=0.98\columnwidth]{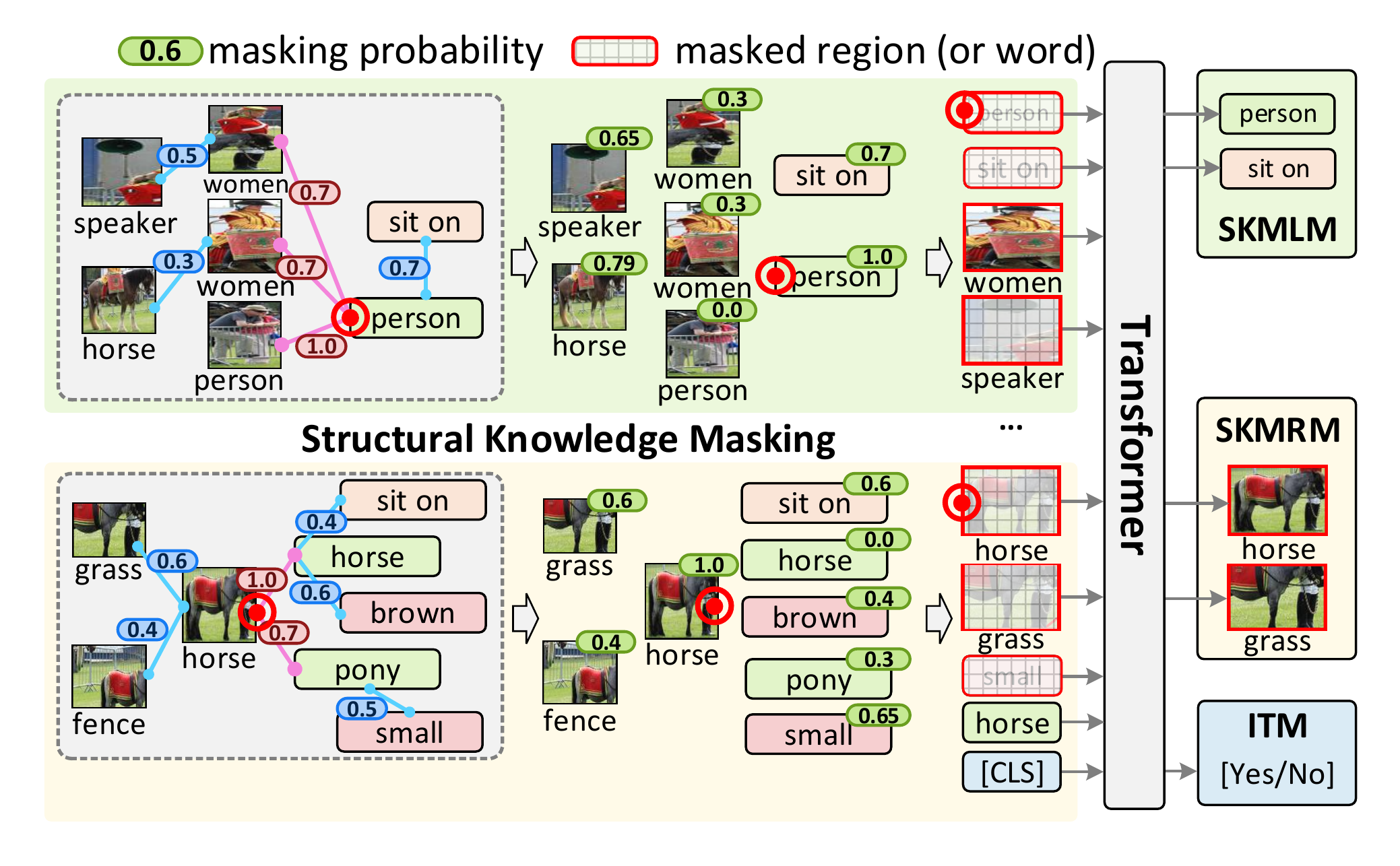}
\vspace{-5pt}
\caption{The flowchart of our ROSITA framework with the structural knowledge masking (SKM) strategy.}
\label{fig:framework}
\end{center}
\vspace{-10pt}
\end{figure}

Based on the extracted knowledge entries from image-text pairs, we introduce the ROSITA framework in this section. We first describe the image and text feature representations and the network architecture. Then, we introduce a structural knowledge masking (SKM) strategy,
which takes the knowledge entries as a priori to perform the masked language (region) modeling. Finally, we describe the whole pretraining objective with multi-task learning. The overall framework is illustrated in Figure \ref{fig:framework}.
\\
\noindent\textbf{Image and Text Feature Representations.} Following the commonly used strategy in existing VLP methods \cite{li2020oscar,chen2015microsoft}, the input image is represented as a set of regional features, which are extracted from a Faster R-CNN model pretrained on Visual Genome \cite{anderson2018bottom}. More specifically, we extract $m$ regions with the highest confidence probabilities from the image. For the $i$-th region, it is represented as a visual feature $f_i\in\mathbb{R}^{2048}$ and a positional feature $p_i\in\mathbb{R}^{5}$ \cite{yu2020ernievil}. The two features are fused into a $d$-dimensional image representation $x_i\in\mathbb{R}^d$ using two linear projections as follows:
\begin{equation}\label{eq:img_rep}
x_i = W_f^Tf_i + W_p^Tp_i
\end{equation}
where $W_f\in\mathbb{R}^{2048 \times d}$ and $W_p\in\mathbb{R}^{5 \times d}$. Finally, the image is represented as a feature matrix $X\in\mathbb{R}^{m\times d}$.

For its paired text, we adopt the word processing method similar to \cite{devlin2019bert}. The input text is first tokenized into words and trimmed (or padded) to a maximum of $n$ words. Each word $w_i$ and its index $i$ (\emph{i.e.}, the absolute position of $w_i$ in the text) are projected to vectors by two individual embedding layers, then added to obtain the position-aware text representation $y_i$ as follows:
\begin{equation}\label{eq:txt_rep}
y_i = \mathrm{WordEmbed}(w_i) + \mathrm{IdxEmbed}(i)
\end{equation}
where $y_i$ is $d$-dimensional to match the image representation. The text is finally represented as a feature matrix  $Y\in\mathbb{R}^{n\times d}$.
\vspace{5pt}
\\
\noindent\textbf{Network Architecture.} The image features $X=[x_1,...,x_m]$ and text features $Y=[y_1,...,y_n]$ are first concatenated before feeding to the network. We insert two special tokens to the concatenated features to obtain the multimodal input features $Z$:
\begin{equation}\label{eq:it_rep}
Z=[x_1,x_2,...,x_m, [\texttt{SEP}], y_1,y_2,...,y_n, [\texttt{CLS}]]
\end{equation}
where the \texttt{[SEP]} token marks the boundary between the image and text features. The \texttt{[CLS]} token is used to predict whether the given image and text are paired or not.

The multimodal features $Z$ are fed into a single-stream Transformer with $L$ layers \cite{devlin2019bert}. Each layer consists of a multi-head self-attention (MSA) block and a feed-forward networks (FFN) block.
\begin{equation}
\begin{aligned}\label{eq:transformer}
\hat{Z}^{\ell} &= \mathrm{LN}(\mathrm{MSA}(Z^{\ell-1})+Z^{\ell-1}),~~~&\ell=1,2,...,L \\
Z^{\ell} &= \mathrm{LN}(\mathrm{FFN}(\hat{Z}^{\ell})+\hat{Z}^{\ell}),~~~&\ell=1,2,...,L \\
\end{aligned}
\end{equation}
where $Z^0=Z$. Layer normalization \cite{ba2016layer} and residual connection \cite{he2016deep} are applied after every block, respectively.
\vspace{5pt}
\\
\noindent\textbf{Structural Knowledge Masking.} The masked language modeling (MLM) \cite{devlin2019bert} and masked region modeling (MRM) \cite{chen2020uniter} tasks are commonly used in almost all the VLP methods \cite{lu2019vilbert,chen2020uniter,yu2020ernievil}. They randomly mask the input tokens (\emph{i.e.}, words or regions) and predict these masked tokens based on their contextual tokens. Since the random masking based MLM and MRM tasks are not aware of the keywords and key regions to be aligned, their efficacy in the alignment learning is weak. To this end, we present an alternative \emph{structural knowledge masking} (SKM) strategy to selectively mask the tokens referred to by the extracted knowledge entry. Accordingly, the MLM and MRM tasks are respectively modified to the SKMLM and SKMRM tasks to adapt to the SKM strategy.

Let an image be represented as a set of regions $\mathcal{R}=\{{r}_1,...,{r}_m\}$ and a text be represented as a sequence of words $\mathcal{W}=\{{w}_1,...,{w}_n\}$, we construct a unified scene graph $G$ on top of $\mathcal{R}$ and $\mathcal{W}$, and extract a set of knowledge entries from $G$. Let $g(v_i)=<\hat{V}, \hat{E}, \hat{S}>$ be one of the knowledge entries, where $v_i\in \hat{V}$ is the anchor object. The vertices are represented as $\hat{V}=\{v_1,...,v_N\}$ and the similarities between the vertices are represented as $\hat{S}\in\mathbb{R}^{N\times N}$, where $N$ is the number of vertices in this entry.

The strategy of SKM is to determinately mask the anchor object $v_i$ while probabilistically masking its intra-modal contexts and cross-modal contexts with respect to the graph structure of the knowledge entry. Since the similarities between $v_i$ and its contexts are different, we assign \emph{independent} masking probabilities to each of the contexts with respect to their similarities to $v_i$, rather than simply using an \emph{identical} masking probability for all the contexts. To obtain the masking probabilities for the contexts, we introduce a masking strategy that satisfies the following principles: for the intra-modal contexts, a larger similarity score refers to a higher masking probability. For the cross-modal contexts, a larger score leads to a lower masking probability. The reasons behind this masking strategy will be explained hereinafter.

Note that not all contexts have direct connections to the anchor object. Therefore, we calculate the transmission probabilities $T=[t_1,...,t_N]\in[0,1]^N$ from the anchor object $v_i$ to its contexts in $\hat{V}$ based on the normalized similarities defined in $\hat{S}$. Since $v_i$ can reach all the vertices in $\hat{V}$ within two hops, $T$ is defined as follows:
\begin{equation}\label{eq:trans_prop}
T=\frac{1}{2}\hat{S}\pi(i)+\frac{1}{2}\hat{S}\hat{S}\pi(i)
\end{equation}
where $\pi(i)\in\{0,1\}^N$ is a one-hot vector with the $i$-th element to be 1. The two terms correspond to the transmission probabilities between the anchor object $v_i$ and its one-hop and two-hop contexts, respectively. After that, we convert the transmission probabilities $T$ to the masking probabilities $P=[p_1,...,p_N]\in[0,1]^N$ using the following rules to satisfy our masking strategy above:
\begin{equation}\label{eq:maskprob}
p_j =
\begin{cases}
    1, & \text{if } v_j \text{ is the anchor object}\\
    \alpha t_j, & \text{if } v_j \text{ is within the intra-modal contexts}\\
    (1-\alpha) (1-t_j), & \text{if } v_j \text{ is within the cross-modal contexts}
\end{cases}
\end{equation}
where $\alpha$ is a hyper-parameter to balance the masking probabilities of the intra-modal and cross-modal contexts. $t_j$ and $p_j$ denote the transmission and masking probability of the vertex $v_j$, respectively.

Given a knowledge entry, we use the calculated masking probabilities to obtain two groups of mask indices $M_w$ and $M_r$, indicating the words and regions to be masked, respectively. The partially masked input features are passed through the network and then fed into the SKMLM and SKMRM tasks.

In particular, if the anchor point $v_i$ refers to an object word in the text, we resort to the SKMLM task to reconstruct the masked words $\mathcal{W}_{{M}_w}$ as follows:
\begin{equation}\label{eq:skmlm}
\mathcal{L}_\mathrm{SKMLM}(\theta) = -\mathbb{E}_{(\mathcal{W}, \mathcal{R})\sim D}~\mathrm{log}~P_\theta(\mathcal{W}_{{M}_w}|\mathcal{W}_{\backslash{M}_w},\mathcal{R}_{\backslash{M}_r})
\end{equation}
where $\theta$ is the trainable parameters. Each pair $(\mathcal{W}, \mathcal{R})$ is sampled from the whole training set $D$. $\mathcal{W}_{\backslash{M}_w}$ and $\mathcal{R}_{\backslash{M}_r}$ refer to the remaining words in $\mathcal{W}$ and the remaining regions in $\mathcal{R}$ with excluding the masked tokens from their modalities, respectively.

Analogously, if $v_i$ refers to a region in the image, we resort to the SKMRM task to reconstruct the masked regions $\mathcal{R}_{{M}_r}$ as follows:
\begin{equation}\label{eq:mrm}
\mathcal{L}_\mathrm{SKMRM}(\theta) = -\mathbb{E}_{(\mathcal{W}, \mathcal{R})\sim D}~f_\theta(\mathcal{R}_{{M}_r}|\mathcal{W}_{\backslash{M}_w},\mathcal{R}_{\backslash{M}_r})
\end{equation}
where $f_{\theta}(\cdot)$ refers to some loss functions. Similar to \cite{chen2020uniter}, we use the regression-based loss and classification-based loss jointly.

The motivations of SKM can be explained as follows: (\emph{i}) The intra-modal contexts may contain interference information (\emph{e.g.}, the word ``\emph{sky}'' is frequently associated with an attribute ``\emph{blue}'' and a visual object of ``\emph{wheel}'' is usually within an object of ``\emph{car}''). Such interference information may leak out the semantics of the anchor object and reduce the difficulty of anchor object reconstruction, leading to a degradation of the pretrained model. Therefore, when masking an anchor object, its intra-modal contexts with high similarities will have high probabilities to be masked simultaneously. This operation reduces the risk of information leakage and enforces the model to acquire precise information from the opposite modality, which \emph{implicitly} enhance the semantic alignments; (\emph{ii}) The cross-modal contexts  with low similarities may contain irrelevant or noisy information. Therefore, when masking an anchor object, its cross-modal contexts with high similarities will have low probabilities to be masked at the same time, which \emph{explicitly} excludes potential noise thus benefiting the semantic alignments. As a result, the synergy of the masking operations above significantly facilitates the semantic alignments.
\vspace{5pt}
\\
\noindent\textbf{Multi-task Learning.} Similar to \cite{chen2020uniter}, we adopt a multi-task learning objective to pretrain our model. Besides the proposed SKMLM and SKMRM tasks, we also include the image-text matching (ITM) task. Moreover, since the SKMLM and SKMRM tasks only focus on the key tokens included in the knowledge entry, we still retain the original random masking-based MLM and MRM tasks to guarantee a good coverage of the remaining tokens in the image and text\footnote{We have made such an experiment that removes the MRM \& MLM tasks. The resulting model reports slight performance drop ($\sim$0.3 points) on the downstream tasks.}.

\section{Experiments}
We evaluate ROSITA on three V+L tasks and perform thorough comparative analysis to the state-of-the-art VLP methods on six datasets. Furthermore, we conduct comprehensive ablation experiments to explore its effectiveness in learning fine-grained semantic alignments.

\subsection{Pretraining Setup}
\noindent\textbf{Datasets.} Following the strategy in \cite{chen2020uniter}, we construct the pretraining dataset consisting of 9.5M train and 155K validation image-text pairs from four existing datasets, namely the COCO Captions \cite{chen2015microsoft}, Visual Genome Captions\cite{krishna2017visual}, Conceptual Captions \cite{sharma2018conceptual}, and SBU Captions \cite{ordonez2011im2text}. The four datasets are categorized into the \emph{in-domain} and \emph{out-of-domain} datasets based on whether they share the same images with the downstream tasks. The statistics of the pretraining dataset are shown in Table \ref{table:corpus}.
\vspace{5pt}
\\
\noindent\textbf{Implementation Details.} For the input image-text pairs, we extract a fixed number of 36 region features from a pre-trained Faster R-CNN model \cite{anderson2018bottom} and adopt the BPE strategy to tokenize the sentence into a maximum of 50 words following \cite{devlin2019bert}. Our ROSITA model adopts a 12-layer Transformer encoder architecture with 768 hidden units and 12 attention heads. The hyper-parameter $\alpha$ in Eq.(\ref{eq:maskprob}) is set to 0.9. The masking probabilities in the original MRM and MLM tasks are set to 15\% \cite{yu2020ernievil}. The model is initialized with the parameters from a pretrained BERT-base model \cite{devlin2019bert}, and then trained up to 40 epochs with a batch size of 512.

\begin{table}
\centering
\caption{The detailed statistics of the used datasets. Following the strategies in \cite{chen2015microsoft}, we split them into in-domain and out-of-domain splits based on the image sources. Each cell shows the number of image-text pairs.}
\label{table:corpus}
\begin{tabular}{c|cc|cc|c}
\toprule
    &  \multicolumn{2}{c|}{in-domain} & \multicolumn{2}{c|}{out-of-domain} &  \multirow{3}{*}{total}\\
\cmidrule{2-5}& \makecell{COCO\cite{chen2015microsoft}} & \makecell{VG\cite{krishna2017visual}} & \makecell {CC\cite{sharma2018conceptual}} & \makecell{SBU \cite{ordonez2011im2text}} &  \\
\midrule
    train & 533K& 5.1M& 3.0M & 869K& 9.5M \\
    val & 25K& 106K& 14K& 10K& 155K\\
\bottomrule
\end{tabular}
\vspace{-5pt}
\end{table}

\begin{table*}
\centering
\caption{Results on \textbf{downstream V+L tasks} to compare with the state-of-the-art VLP methods. For a fair comparison, all the results are archived by the base models. Most of the models are trained on the \emph{in-domain}+\emph{out-of-domain} datasets, except for those models marked with $\dag$ are trained on the \emph{out-of-domain} datasets. IR and TR denote the image retrieval and text Retrieval, respectively. For the REC task, all the results are achieved based on the detected region features from images. Dark and light grey colors highlight the top and second best results on each evaluation metric.}\label{table:mainres}
\vspace{-5pt}
\begin{tabular}{c|cl|cccccccc|c}
\toprule
task
 & \multicolumn{2}{c|}{dataset}
 &  \makecell{ViLBERT$^\dag$\\\cite{lu2019vilbert}}
 & \makecell{VLBERT$^\dag$\\\cite{su2019vl}}
 & \makecell{Unicoder-VL\\\cite{li2020unicoder}}
 & \makecell{LXMERT\\\cite{tan2019lxmert}}
 & \makecell{UNITER\\\cite{chen2020uniter}}
 & \makecell{ERNIE-ViL$^\dag$\\\cite{yu2020ernievil}}
 & \makecell{VILLA\\\cite{gan2020large}}
 & \makecell{OSCAR\\\cite{li2020oscar}}
 & \makecell{ROSITA\\(ours)}\\
\midrule
\multirow{2}{*}{VQA}& \multirow{2}{*}{VQAv2} & test-dev & 70.55 & 71.16 & - & 72.42 & 72.70 & 72.62 & \multicolumn{1}{>{\columncolor{mygray1}}c}{73.59} & 73.16 &  \multicolumn{1}{>{\columncolor{mygray2}}c}{73.91} \\
 & & test-std & 70.92 & - & - & 72.54 & 72.91 & 72.85 & \multicolumn{1}{>{\columncolor{mygray1}}c}{73.67} & 73.44 &  \multicolumn{1}{>{\columncolor{mygray2}}c}{73.97}\\
    \midrule
\multirow{8}{*}{REC}&  \multirow{3}{*}{\makecell{Ref-\\COCO}}  & val$^d$ & - & - & - & - & 81.24 & - & \multicolumn{1}{>{\columncolor{mygray1}}c}{81.65} & - & \multicolumn{1}{>{\columncolor{mygray2}}c}{84.79}\\
&  & testA$^d$ & - & - & - & - & 86.48 & - & \multicolumn{1}{>{\columncolor{mygray1}}c}{87.40} & - & \multicolumn{1}{>{\columncolor{mygray2}}c}{87.99}\\
&  & testB$^d$ & - & - & - & - & 73.94 & - & \multicolumn{1}{>{\columncolor{mygray1}}c}{74.48}& -& \multicolumn{1}{>{\columncolor{mygray2}}c}{78.28}\\
    \cmidrule{2-12}
& \multirow{3}{*}{\makecell{Ref-\\COCO+}}  & val$^d$ & 72.34 & 71.60 & - & - & 75.31 & {74.02} & \multicolumn{1}{>{\columncolor{mygray1}}c}{76.05} & - & \multicolumn{1}{>{\columncolor{mygray2}}c}{76.06}\\
&  & testA$^d$ & 78.52 & 77.72 & - & - & 81.30 & {80.33} & \multicolumn{1}{>{\columncolor{mygray1}}c}{81.65} & - & \multicolumn{1}{>{\columncolor{mygray2}}c}{82.01}\\
&  & testB$^d$ & 62.61 & 60.99 & - & - & 65.68 & {64.74} & \multicolumn{1}{>{\columncolor{mygray1}}c}{65.70} & - & \multicolumn{1}{>{\columncolor{mygray2}}c}{67.40}\\
   \cmidrule{2-12}
& \multirow{2}{*}{\makecell{Ref-\\COCOg}}  & val$^d$ & - & - & - & - & 74.31 & - & \multicolumn{1}{>{\columncolor{mygray1}}c}{75.90} & - & \multicolumn{1}{>{\columncolor{mygray2}}c}{78.23}\\
&  & test$^d$ & - & - & - & - & 74.51 & - & \multicolumn{1}{>{\columncolor{mygray1}}c}{75.93} & - & \multicolumn{1}{>{\columncolor{mygray2}}c}{78.25}\\
   \midrule
\multirow{12}{*}{ITR}& \multirow{3}{*}{\makecell{IR-\\COCO}} & R@1 & - & - & 46.70 & - & 50.33 & - & - & \multicolumn{1}{>{\columncolor{mygray1}}c|}{54.00} & \multicolumn{1}{>{\columncolor{mygray2}}c}{54.40}\\
&  & R@5 & - & - & 76.00 & - & 78.52 & - & - & \multicolumn{1}{>{\columncolor{mygray1}}c|}{80.80} & \multicolumn{1}{>{\columncolor{mygray2}}c}{80.92}\\
&  & R@10 & - & - & 85.30 & - & 87.16 & - & - & \multicolumn{1}{>{\columncolor{mygray1}}c|}{88.50} & \multicolumn{1}{>{\columncolor{mygray2}}c}{88.60}\\
    \cmidrule{2-12}
& \multirow{3}{*}{\makecell{TR-\\COCO}} & R@1 & - & - & 62.30 & - & 64.40 & - & - & \multicolumn{1}{>{\columncolor{mygray1}}c|}{70.00} & \multicolumn{1}{>{\columncolor{mygray2}}c}{71.26}\\
&  & R@5 & - & - & 87.10 & - & 87.40 & - & - & \multicolumn{1}{>{\columncolor{mygray1}}c|}{91.10} & \multicolumn{1}{>{\columncolor{mygray2}}c}{91.62}\\
&  & R@10 & - & - & 92.80 & - & 93.08 & - & - & \multicolumn{1}{>{\columncolor{mygray1}}c|}{95.50} & \multicolumn{1}{>{\columncolor{mygray2}}c}{95.58}\\
   \cmidrule{2-12}
& \multirow{3}{*}{\makecell{IR-\\Flickr}}  & R@1 & 58.20 & - & 71.50 & - & 72.52 & \multicolumn{1}{>{\columncolor{mygray2}}c}{74.44} & \multicolumn{1}{>{\columncolor{mygray1}}c}{74.74} & - & {74.08}\\
&  & R@5 & 84.90 & - & 90.90 & - & 92.36 & \multicolumn{1}{>{\columncolor{mygray1}}c}{92.72} & \multicolumn{1}{>{\columncolor{mygray2}}c}{92.86} & - & {92.44}\\
&  & R@10 & 91.52 & - & 94.90 & - & \multicolumn{1}{>{\columncolor{mygray2}}c}{96.08} & 95.94 & 95.82 & - & \multicolumn{1}{>{\columncolor{mygray2}}c}{96.08}\\
    \cmidrule{2-12}
&\multirow{3}{*}{\makecell{TR-\\Flickr}}  & R@1 & - & - & 86.20 & - & 85.90 & \multicolumn{1}{>{\columncolor{mygray1}}c}{86.70} & 86.60 & - & \multicolumn{1}{>{\columncolor{mygray2}}c}{88.90}\\
&  & R@5 & - & - & 96.30 & - & 97.10 & 97.80 & \multicolumn{1}{>{\columncolor{mygray1}}c}{97.90} & - & \multicolumn{1}{>{\columncolor{mygray2}}c}{98.10}\\
&  & R@10 & - & - & 99.00 & - & 98.80 & 99.00 & \multicolumn{1}{>{\columncolor{mygray1}}c}{99.20} & - & \multicolumn{1}{>{\columncolor{mygray2}}c}{99.30}\\
\bottomrule
\end{tabular}
\vspace{-5pt}
\end{table*}

\subsection{Downstream Tasks}
After obtaining the pretrained ROSITA model, we finetune it on three downstream V+L tasks as follows.
\vspace{5pt}
\\
\noindent\textbf{Visual Question Answering (VQA)} is a task that requires the model to answer natural language questions about an image. We adopt the widely used VQAv2 dataset \cite{antol2015vqa,goyal2017making}, which is manually built on the images from the MSCOCO dataset \cite{lin2014microsoft}. The dataset is split into train (83k images and 444k questions), validation (41k images and 214k questions), and test (81k images and 448k questions) sets. Following the strategy in \cite{chen2020uniter}, we feed the representation of the $\mathsf{[CLS]}$ token to a linear classifier to predict the corresponding answer from a vocabulary of size 3129 \cite{yu2019deep}.
\vspace{5pt}
\\
\noindent\textbf{Referring Expression Comprehension (REC)} is a task that  requires to localize an image region referred to by a natural language query. We evaluate the performance on RefCOCO \cite{kazemzadeh2014referitgame}, RefCOCO+ \cite{kazemzadeh2014referitgame} and RefCOCOg \cite{mao2016generation} datasets. All the three datasets are collected from COCO images [31]. RefCOCO and RefCOCO+ are split into four subsets, including train (120k queries), validation (11k queries), testA (6k queries about people), and testB (6k queries about objects), while RefCOCOg is split into three subsets, including train (81k queries), validation (5k queries), and test (10k queries). The representation for each image region is used to predict a ranking score and a refined bounding box.
\vspace{5pt}
\\
\noindent\textbf{Image-Text Retrieval (ITR)} is a task that requires the model to calculate a similarity score between an image and a sentence and then perform cross-modal retrieval. We conduct experiments on the COCO Captions \cite{chen2015microsoft} and Flickr30K \cite{young2014image} datasets, respectively. Following the partition strategy by \cite{karpathy2015deep}, the COCO dataset is split into 82k/5k/5k train/validation/test images, while the Flickr30K dataset is split into 29k/1k/1k train/validation/test images. Similar to \cite{chen2020uniter}, we use an offline hard sample mining strategy to obtain 128 negative samples per each positive sample, and use the representation of the $\mathsf{[CLS]}$ token to predict a matching score.

\begin{table*}
\centering
\caption{Ablations of ROSITA variants without the cross- and intra-modal knowledge. All models are pretrained on the \emph{in-domain} datasets and then finetuned on specific downstream tasks. For each model, we report the accuracies on the pretraining tasks an downstream tasks, respectively. As we only have positive image-text pairs in the pretraining datasets, we use the offline hard sample mining strategy to generate an equal number of negative samples for the evaluation of the ITM task.}\label{table:aba_know}
\vspace{-5pt}
		\begin{tabular}{cl|ccc|cccc}
            \toprule
         \multirow{4}{*}{\#}   & \multirow{4}{*}{model} & \multicolumn{3}{c|}{pretraining tasks} & \multicolumn{4}{c}{downstream tasks}\\
            \cmidrule{3-9}
            && ITM& SKMLM & SKMRM & \makecell{VQAv2\\(dev)} & \makecell{RefCOCO\\(val)} & \makecell{IR-Flickr\\(test)} & \makecell{TR-Flickr\\(test)} \\
            \midrule
            1 & ROSITA (full)  &\textbf{84.34}&\textbf{67.16}&\textbf{76.50}&  \textbf{73.19} &  \textbf{84.22} &   \textbf{85.09}   & \textbf{94.33} \\
             2 & -w/o cross-modal  knowledge &83.54&63.69&72.56&   72.86&  83.85&   84.23    & 93.63 \\
             3 &  -w/o intra-modal knowledge  &83.30&63.75&73.90& 72.98  &  83.31 & 84.79  & 93.90  \\
             4 &  -w/o both types of knowledge   &82.22& 61.19&68.58& 72.47  & 82.12  &   82.11  & 92.57\\
            \bottomrule
        \end{tabular}
\vspace{-5pt}
\end{table*}

\subsection{Main Results}
We compare the proposed ROSITA model against existing state-of-the-art VLP methods. As shown in Table \ref{table:mainres}, ROSITA achieves the overall best performance on all downstream tasks, which verifies the effectiveness of the integrated cross- and intra-modal knowledge and the corresponding SKM strategy\footnote{We have conduct such an experiment that pretrains ROSITA on the \emph{out-of-domain} datasets only. The resulting model consistently outperforms the counterparts \cite{yu2020ernievil,lu2019vilbert,su2019vl}, verifying the generalization capability of our approach.}.

It is worth noting that some methods like ViLBERT, LXMERT, and ERNIE-ViL adopt the two-stream architecture, which have much more parameters (ROSITA: 116M, VilBERT: 221M, LXMERT: 183M, ERNIE-ViL: 228M). Some methods like UNITER and VILLA use a larger number of image features (up to 100 regions), which has been verified to benefit the performance at the expense of much higher computational cost. In contrast, ROSITA uses a fixed number of 36 image features. We believe the performance of ROSITA can be further improved by taking these advanced strategies above.

\subsection{Ablation Studies}
\begin{figure*}
\begin{center}
\includegraphics[width=0.98\textwidth]{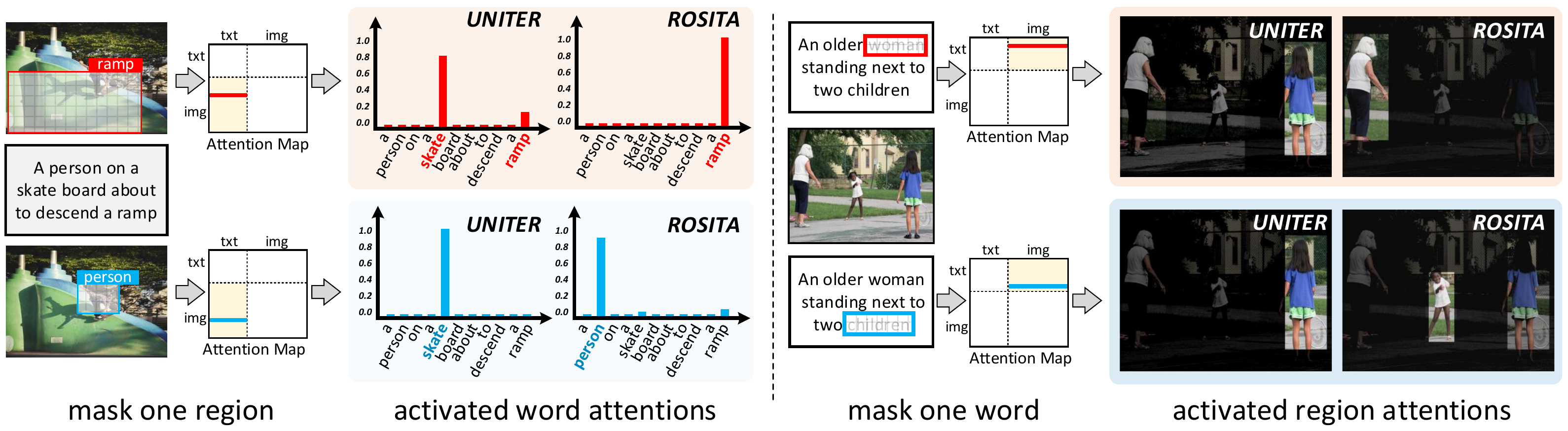}
\vspace{-5pt}
\caption{Visualizations of the learned cross-modal attentions (\emph{i.e.}, region-to-words attentions on the left and word-to-regions on the right) from UNITER \cite{chen2020uniter} and ROSITA. Taking the image-text pair as inputs with exactly one region (or word) being masked at a time, we extract the attention map from the last MSA block of the pretrained model. The region-to-words (word-to-regions) attentions correspond to one specific row in the bottom-left (top-right) area of the attention map, respectively.}
\label{fig:attvis}
\vspace{-10pt}
\end{center}
\end{figure*}

\begin{figure*}
\begin{center}
\includegraphics[width=0.98\textwidth]{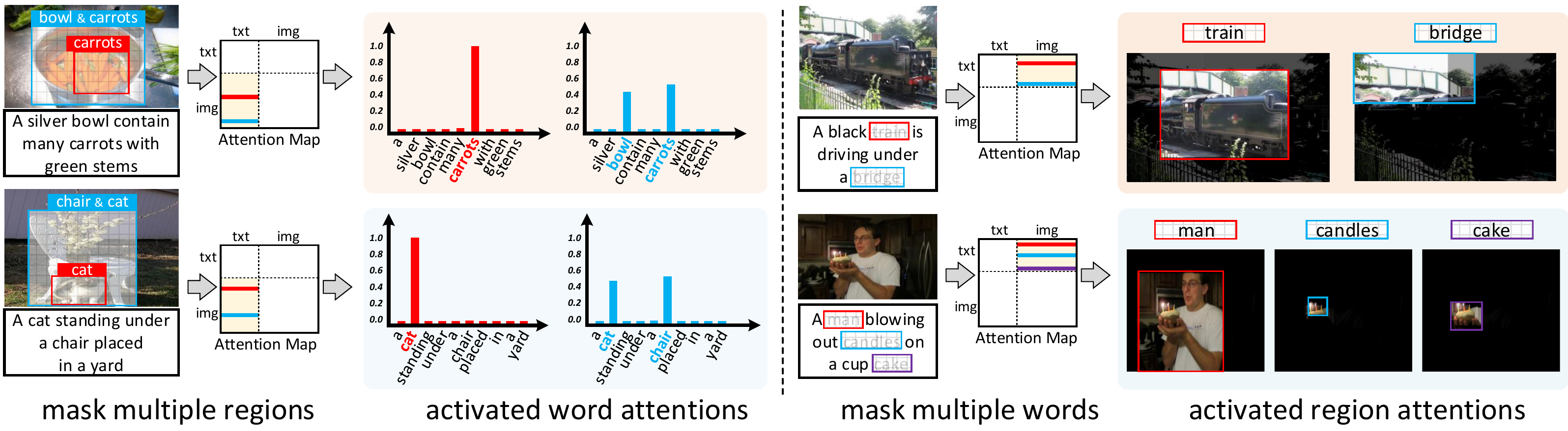}
\vspace{-5pt}
\caption{Visualizations of the region-to-words attentions (left) and word-to-regions attentions (right) from a pretrained ROSITA model with masking multiple regions (or words) at one time.}
\label{fig:comp_attvis}
\vspace{-5pt}
\end{center}
\end{figure*}

We run a number of ablations to investigate the reasons of ROSITA's effectiveness.
The results show in Table \ref{table:aba_know}-\ref{table:aba_skm} and Figure \ref{fig:attvis}-\ref{fig:comp_attvis} are discussed in detail below.
\begin{table}
\centering
\caption{Ablations of four ROSITA variants with two alternative masking strategy in SKM (\emph{i.e.}, independent probabilities and identical probability). All models are pretrained on the \emph{in-domain} datasets and finetuned on the downstream tasks.}\label{table:aba_skm}
\vspace{-5pt}
		 \begin{tabular}{c|cccc}
            \toprule
             \multirow{2}{*}{masking prob.} &  VQAv2 & {RefCOCO} & {IR-Flickr} & {TR-Flickr} \\
              & (dev) & (val)  & (test) & (test) \\
            \midrule
          independent&  \textbf{73.19} &  \textbf{84.22} &   \textbf{85.09}   & \textbf{94.33} \\
            identical ($p$=45\%)& 72.79& 83.18 &  83.70 & 93.20\\
            identical ($p$=30\%)&72.93 &    83.29 & 84.36 & 93.63\\
            identical  ($p$=15\%)& 72.75&  82.96 & 83.75 & 93.53\\
            \bottomrule
        \end{tabular}
        \vspace{-10pt}
\end{table}
\vspace{5pt}
\\
\noindent\textbf{Cross- and Intra-modal Knowledge.} In Table \ref{table:aba_know}, we show the effects of the intra-modal knowledge and cross-modal knowledge based on the performance on the pretraining and downstream tasks. Taking the full ROSITA as the reference model (Line \#1), we obtain the different variants by removing the cross-modal knowledge or the intra-modal knowledge. The variant without cross-modal knowledge (Line \#2) indicates that the model is not aware of the anchor objects and the SKM strategy is performed only on a single modality using the intra-modal knowledge. In contrast, the variant without intra-modal knowledge (Line \#3) indicates that the model is aware of the anchor objects but is not aware of the intra-modal contexts. Finally, by removing both the cross- and intra-modal knowledge, we obtain a baseline variant nearly identical to UNITER \cite{chen2020uniter} (Line \#4)\footnote{Our model has slight performance deviations compared with the original UNITER model since we use different visual features and pretraining hyper-parameters.}.

Given the pretrained models of the four variants above (\emph{i.e.}, without finetuning on downstream tasks), we evaluate their performance on three pretraining tasks. The ITM task examines the ability of semantic alignment between image-text pairs. From the results, we can see that both types of knowledge bring performance improvement to the ITM task (\#4 \emph{vs.} \#3 and \#2). Moreover, the two types of knowledge are complementary that their synergy brings 2.1 points improvement compared to the baseline model without any knowledge (\#4 \emph{vs.} \#1). Although the ITM task is the most straightforward metric for semantic alignment, it only measures the \emph{coarse-grained} alignments on the image-text level, thus cannot fully reveal the capability of ROSITA. As a complement, we resort to the SKMLM and SKMRM tasks to evaluate the \emph{fine-grained} alignments on the region-word level. Compared with the baseline model in \#4, the full ROSITA model improves the accuracies by 7.0 and 7.9 points on the SKMLM and SKMRM tasks, respectively.

Next, we report the performance of these variants on different downstream tasks. From the demonstrated results, we obtain similar observations to those on the pretraining tasks. The full ROSITA model consistently outperforms all the counterparts, verifying the effectiveness of the cross- and intra-modal knowledge.
\vspace{5pt}
\\
\noindent\textbf{SKM Strategy.} After extracting knowledge entries from image-text pairs, we have two alternative masking strategies in SKM, \emph{i.e.}, the independent probabilities and the identical probability. For the masking strategy with identical probability, we evaluate the choices of different probabilities within $\{15\%, 30\%, 45\%\}$. The results in Table \ref{table:aba_skm} show that the model pretrained with independent probabilities steadily outperforms all the counterparts with the identical probability. For the models pretrained with the identical probability strategy, their performance is sensitive to the choices of the predefined probability. A small masking probability (\emph{e.g.}, 15\%) may degrade the model towards the baseline without any knowledge. A large masking probability (\emph{e.g.}, 45\%) may shield the essential information that is necessary to learn the semantic alignments. In comparison, the masking strategy with independent probabilities provides a more fine-grained understanding of the knowledge structure, leading to a more robust pretrained model.
\vspace{5pt}
\\
\noindent\textbf{Cross-modal Semantic Alignments.} The effect of \emph{fine-grained} semantic alignments across modalities can be inferred from the attention maps of the learned Transformer model \cite{cao2020behind}. We visualize the learned \emph{cross-modal attentions} (\emph{i.e.}, region-to-words and word-to-regions attentions) from the pretrained UNITER \cite{chen2020uniter} and our ROSITA models, as shown in Figure \ref{fig:attvis}. Taking the image-text pair as inputs with exactly one token (a region or a word) being masked at a time, we pass the multimodal features through the pretrained model and extract the attention map from the last MSA block\footnote{We perform element-wise addition over the attention maps from different heads followed by row-wise softmax normalization to obtain one aggregated attention map.}. The region-to-words and word-to-regions attentions of the masked token correspond to one specific row in the bottom-left and top-right area of the attention map, respectively.

From the visualized cross-modal attentions, we can see that ROSITA learns significantly better semantic alignments than UNITER. ROSITA can precisely align the masked object to its reference object in the opposite modality while UNITER fails to establish such cross-modal alignments. For example, when the region of ``\emph{ramp}'' is masked, ROSITA activates the word ``\emph{ramp}'' precisely while UNITER obtains the largest attention value on the word ``\emph{skate}''. When another region of ``\emph{person}'' is masked, ROSITA precisely activates the word ``\emph{person}'' while UNITER still activates the incorrect word ``\emph{skate}''. Similar phenomena are observed in the opposite direction. ROSITA activates the accurate regions to the masked words while UNITER fails to do it.

To step further, we conduct a more challenging task as follows. We mask \emph{multiple} regions (or words) at the same time to examine whether the semantic alignments can still be achieved. The visualized results in Figure \ref{fig:comp_attvis} show that ROSITA works surprisingly well to establish accurate semantic alignment for each masked token. For example, when the regions of ``\emph{bowl}'' and ``\emph{bowl \& carrots}'' are masked simultaneously, the region of ``\emph{carrot}'' is precisely aligned to the word ``\emph{carrots}'', and the region of ``\emph{bowl \& carrots}'' is aligned to the two words ``\emph{bowl}'' and ``\emph{carrots}'' uniformly.
In the opposite direction, when the words ``\emph{man}'', ``\emph{candles}'', and ``\emph{cake}'' are masked at the same time, their corresponding regions are highlighted in the learned attentions, respectively.


\section{Conclusion}
In this paper, we present a new VLP method called ROSITA, which integrates the cross- and intra-modal knowledge in a unified scene graph to enhance the learning of cross-modal semantic alignment. We introduce a novel structural knowledge masking (SKM) strategy to perform masked language (region) modeling with respect to the knowledge entries extracted from the unified scene graph. Extensive ablations, comparative experiments, and comprehensive analysis show that ROSITA significantly outperforms existing state-of-the-art VLP approaches on three typical V+L tasks over six benchmark datasets. We hope our study will be helpful to inspire future research in the vision-and-language community and beyond.

\begin{acks}
This work was supported in part by the National Key R\&D Program of China under Grant 2018AAA0100603, and in part by National Natural Science Foundation of China under Grant 62072147, Grant 61836002, and Grant 62020106007.
\end{acks}

\newpage
\bibliographystyle{ACM-Reference-Format}
\bibliography{mfp0493}


\end{document}